%% file: ContFormer.tex
\documentclass[10pt,twocolumn,letterpaper]{article}

\usepackage[pagenumbers]{cvpr} 

\usepackage{graphicx}
\usepackage{amsmath}
\usepackage{amssymb}
\usepackage{booktabs}
\usepackage{comment}
\usepackage{multirow}
\usepackage{caption}

\usepackage{graphicx}
\usepackage{color}
\usepackage{multirow}
\usepackage{amsmath,amsfonts}
\usepackage{array}
\newcolumntype{x}[1]{>{\centering\arraybackslash}p{#1pt}}

\usepackage{enumitem}
\usepackage{bm}
\usepackage{xcolor}
\usepackage{booktabs}
\usepackage{multirow}
\usepackage{arydshln}
\usepackage{pifont}

\newcommand\blfootnote[1]{%
	\begingroup
	\renewcommand\thefootnote{}\footnote{#1}%
	\addtocounter{footnote}{-1}%
	\endgroup
}

\usepackage[pagebackref,breaklinks,colorlinks]{hyperref}

\usepackage[capitalize]{cleveref}
\crefname{section}{Sec.}{Secs.}
\Crefname{section}{Section}{Sections}
\Crefname{table}{Table}{Tables}
\crefname{table}{Tab.}{Tabs.}

\def\cvprPaperID{*****} 
\def\confName{CVPR}
\def\confYear{2023}

\begin{document}

\title{Video Referring Expression Comprehension via Transformer \\with Content-aware Query}

\author{Ji Jiang$^{1*}$, Meng Cao$^{1*}$, Tengtao Song$^{1}$, Yuexian Zou$^{\dagger1,2}$\\
	$^{1}$School of Electronic and Computer Engineering, Peking University $^{2}$Peng Cheng Laboratory\\
}
\maketitle

\begin{abstract}
Video Referring Expression Comprehension (REC) aims to localize a target object in video frames referred by the natural language expression. Recently, the Transformer-based methods have greatly boosted the performance limit. However, we argue that the current \emph{query} design is sub-optima and suffers from two drawbacks: 1) the slow training convergence process; 2) the lack of fine-grained alignment. To alleviate this, we aim to couple the pure learnable queries with the content information. Specifically, we set up a fixed number of learnable bounding boxes across the frame and the aligned region features are employed to provide fruitful clues. Besides, we explicitly link certain phrases in the sentence to the semantically relevant visual areas. To this end, we introduce two new datasets (\ie, VID-Entity and VidSTG-Entity) by augmenting the VID-Sentence and VidSTG datasets with the explicitly referred words in the whole sentence, respectively. Benefiting from this, we conduct the fine-grained cross-modal alignment at the \emph{region-phrase} level, which ensures more detailed feature representations. Incorporating these two designs, our proposed model (dubbed as \textbf{ContFormer}) achieves the state-of-the-art performance on widely benchmarked datasets. For example on VID-Entity dataset, compared to the previous SOTA, ContFormer achieves 8.75\% absolute improvement on Accu.@0.6. The dataset, code and models are available at \url{https://github.com/mengcaopku/ContFormer}.
\end{abstract}
\blfootnote{$*$ denotes the equal contributions. $\dagger$ denotes the corresponding author.}

\section{Introduction} \label{sec:intro}
Referring Expression Comprehension (REC)~\cite{hu2017modeling,hu2016natural,yu2016modeling,yu2017joint} aims to locate the image region described by the natural language query. This task has attracted extensive attention from both academia and industry, due to its wide application, such as visual question answering~\cite{antol2015vqa,ji2022visual}, image/video analysis~\cite{anderson2018bottom,cao2022deep,cao2022locvtp} and relationship modeling~\cite{hu2017modeling,zhang2021synergic}. During the past years, most previous works restrict REC in static images~\cite{wang2019neighbourhood,yang2019cross,yu2018mattnet,liao2020real,yang2019fast,mao2016generation}. Recently, with the increasing number of videos uploaded online, grounding the target object in the video is becoming an emerging requirement and some recent attempts~\cite{zhou2018weakly,vasudevan2018object,chen2019weakly,zhang2020does,feng2021siamese} begin  to conduct REC in the video domain. Different from image REC, video REC is more challenging since it needs to deal with both complex temporal and spatial information.

\begin{figure}[t]
	\centering
	\includegraphics[width=0.4\textwidth]{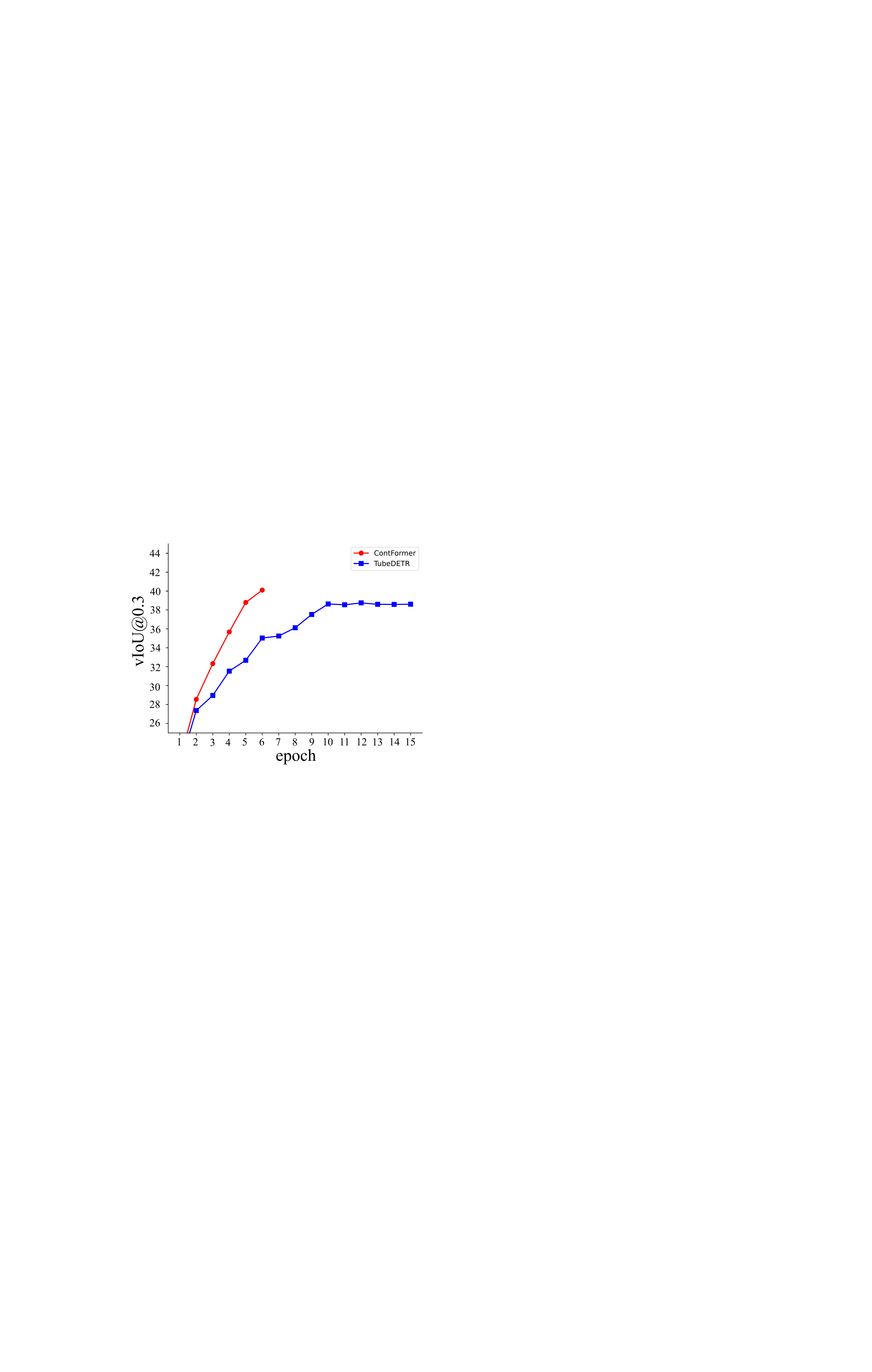}
	\vspace{-0.5em}
	\caption{Comparisons of the convergence curves between TubeDETR and our ContFormer.}
	\label{fig:TubeDETR_ContFormer}
\end{figure}

Current video REC methods can be classified into two major categories: \emph{two-stage}, \emph{proposal-driven} methods and \emph{one-stage}, \emph{proposal-free} methods. For the two-stage methods~\cite{zhang2020does,feng2021decoupled,huang2018finding,gao2017tall}, they extract potential spatio-temporal tubes and then align these candidates to the sentence to find the best matching one. The other stream of one-stage methods~\cite{song2021co,sadhu2020video,zeng2020dense,chen2021end,cao2022correspondence} fuses visual-text features and directly predicts bounding boxes densely at all spatial locations. These two kinds of methods, however, are time-consuming since they require some post-processing steps (\eg, non-maximum suppression, NMS). Recently, DETR-like methods~\cite{carion2020end} have been demonstrated effective in object detection areas, which get rid of the manually-designed rules and dataset-depend hyper-parameters. Following this pipeline, the primary work TubeDETR~\cite{yang2022tubedetr} develops a similar transformer model for video REC.

\begin{figure}[t]
	\centering
	\includegraphics[width=0.4\textwidth]{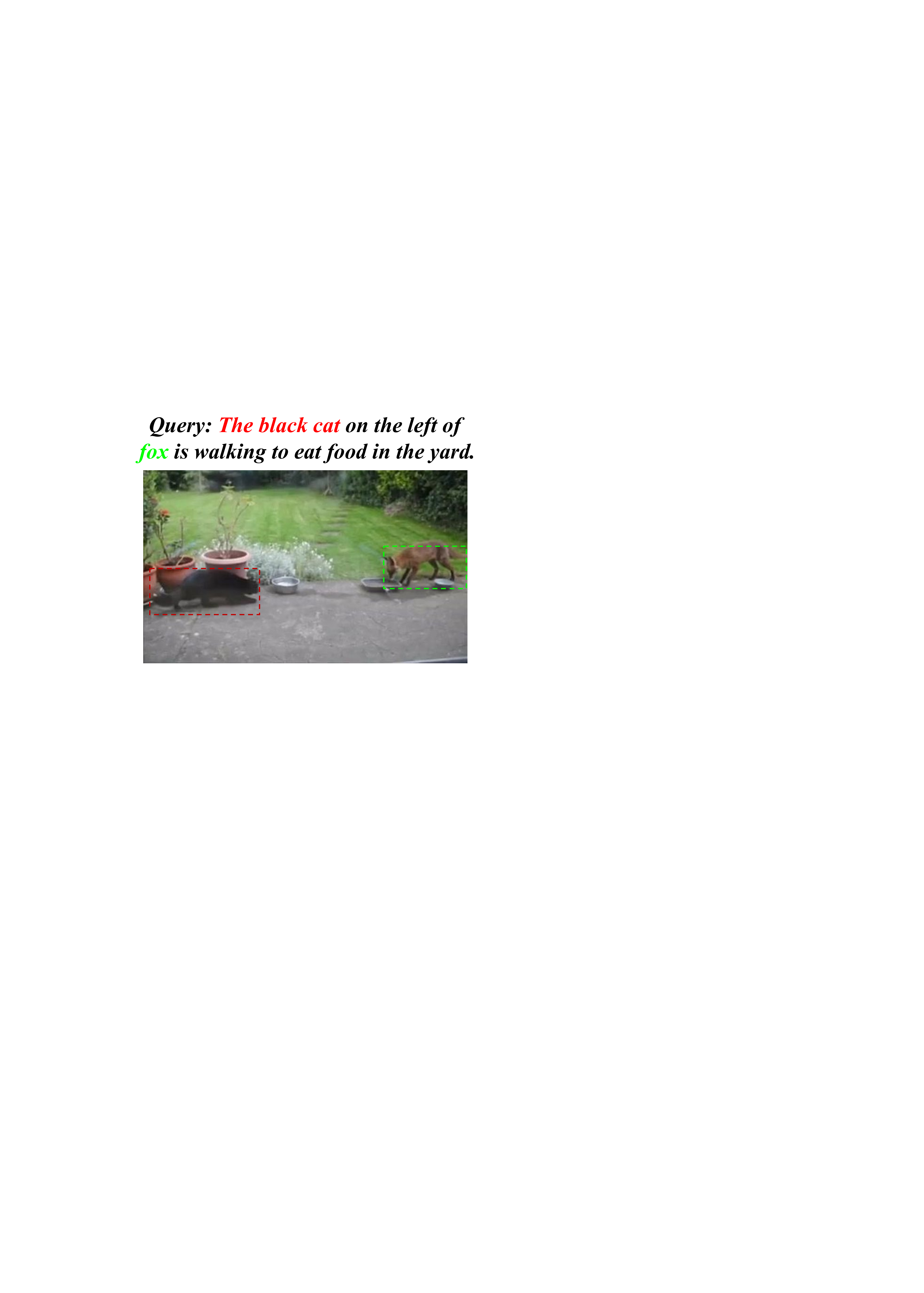}
	\vspace{-0.5em}
	\caption{The certain \emph{regions} (\ie, object areas) of the frame are usually more salient and highly overlapped with certain \emph{phrases} containing semantic meanings.}
	\label{fig:align_example}
\end{figure}

Although noticeable improvements have been achieved, this vanilla method still has two overlooked drawbacks: \emph{1) The slow training convergence process.} DETR-like methods formulate detection/localization as a set prediction problem and use learnable \emph{queries} to probe and pool frame features. This structure, however, suffers from the notorious \emph{slow convergence} issue~\cite{liu2022dab,meng2021conditional,wang2021anchor}. For example in Figure~\ref{fig:TubeDETR_ContFormer}, TubeDETR requires about 10 epochs to achieve the saturated performance. Such a problem greatly hinders its practical applications. \emph{2) Lack of fine-grained alignments.} Empirically, we find that the noun (\ie, subject or object) in a sentence is important to carry the overall meaning. Accordingly, certain patches (\ie, object areas) of the frame are usually more salient and highly overlapped with the semantic meanings. For example in Figure~\ref{fig:align_example}, the sentence contains two instances, \ie, ``\emph{cat}" and ``\emph{fox}". The detailed alignment and differentiation between the mentioned query objects and the corresponding visual areas provide localization clues. This fine-grained correlation, however, is overlooked in current Transformer-based methods. 

Based on the above observations, we argue that the current query design in video REC methods is sub-optimal. To alleviate this, we propose the novel \textbf{cont}ent-aware query in trans\textbf{former} (dubbed as \textbf{ContFormer}). We contend that the content-independent query design is the main cause of the slow convergence. To this end, we propose to use query embeddings conditioned on the image content. Specifically, we set up a fixed number of bounding boxes across the frame. Then the cropped and pooled regional features are transformed into the query features of Transformer decoder. Compared to the conventional high-dimension learnable queries, our region-based features introduce more salient prior, leading to a faster convergence process (cf. Figure~\ref{fig:TubeDETR_ContFormer}).

Besides, current datasets only contain the coarse-grained \emph{region-sentence} level correspondences. In this work, we take one step further to collect VID-Entity and VidSTG-Entity datasets (cf. Figure~\ref{fig:dataset_example}), which annotate  \emph{region-phrase} lables by grounding specific phrases in sentences with the bounding boxes in the video frames. To further use these detailed annotations, we also propose a fine-grained alignment loss. Specifically, we firstly compute the similarity scores between each query-word pair. Then, we adopt the Hungarian algorithm~\cite{kuhn1955hungarian} to select the query matching the target bounding box. Supervised by the annotations of VID-Entity and VidSTG-Entity datasets, the InfoNCE loss is applied to map the fine-grained matched pair to be close.

We make three contributions in this paper:

\begin{itemize}[topsep=0pt, partopsep=0pt, leftmargin=13pt, parsep=0pt, itemsep=3pt]
	\item We contend that the current query design leads to the slow convergence process in Transformer-based video REC methods. To this end, we propose to generate content-conditioned queries based on the frame context.
	\item Beyond the coarse-grained region-sentence one, we build two datasets (\ie, VID--Entity and VidSTG--Entity) and a fine-grained alignment loss to enhance the fine-grained \emph{region-phrase} alignment.
	\item Experimental results show that our ContFormer achieves state-of-the-art performance on both trimmed and untrimmed video REC benchmarks. 
\end{itemize}

\begin{figure*}[t]
	\centering
	\includegraphics[width=0.9\textwidth]{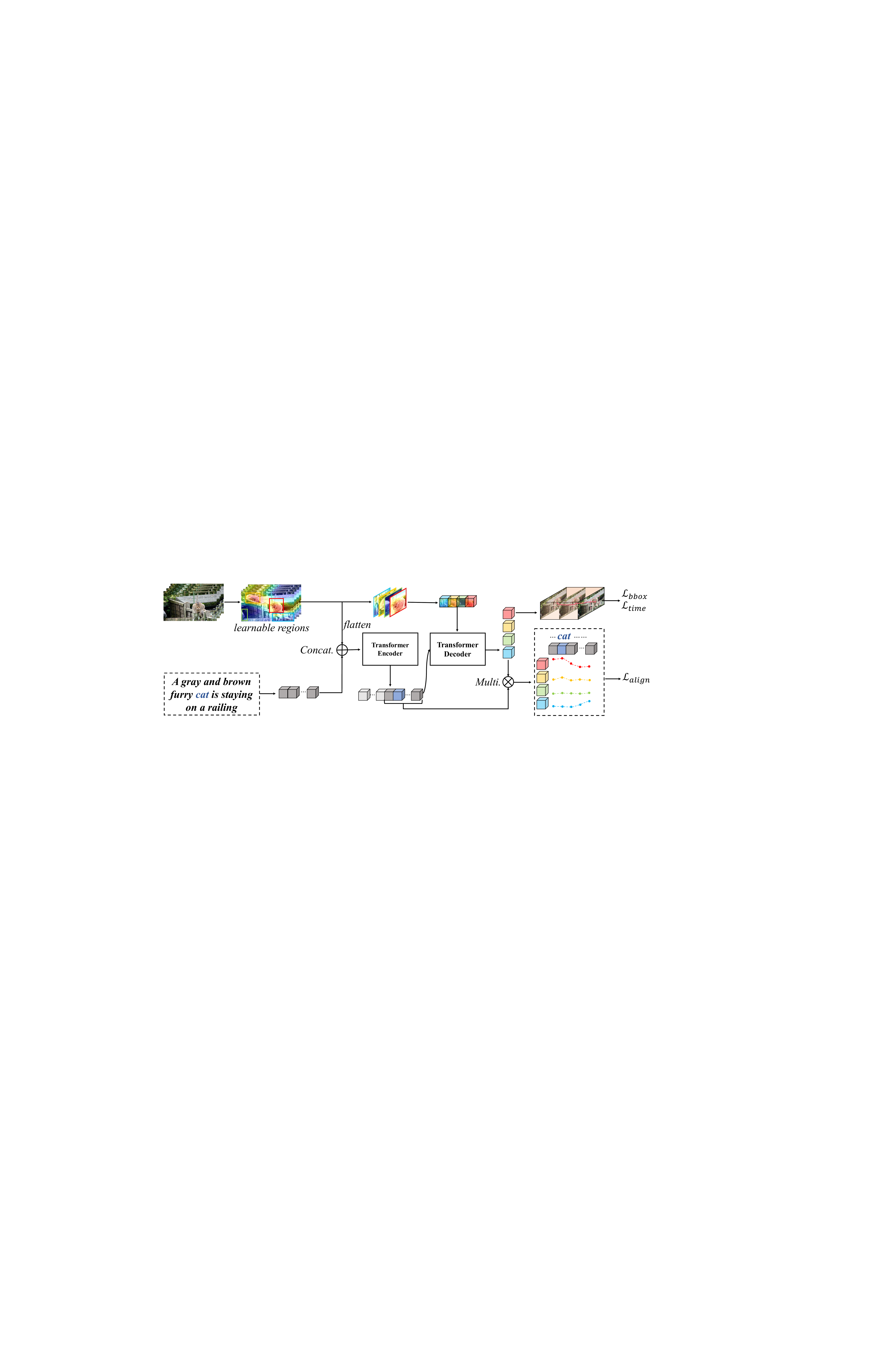}
	\vspace{-0.5em}
	\caption{The schematic illustration of our ContFormer. The video and text modalities are extracted by the modality-specific backbones and fused through the Transformer encoder. A novel content-aware query generation module is proposed for the Transformer decoder to generate content-conditioned query feature. The overall pipeline is optimized by the bi-partial matching loss and our proposed entity-aware contrastive loss.}
	\label{fig:ContFormer_pipeline}
\end{figure*}

\section{Related Work}
\noindent \textbf{Video Referring Expression Comprehension.} 
The objective of video REC is to localize the spatial-temporal tube according to the natural language query. Most of the previous works~\cite{zhang2020does,feng2021decoupled,song2021co,sadhu2020video,hong2020weakly} can be divided into two categories, \ie, two-stage methods and one-stage methods. However, both kinds of methods require time-consuming post-processing steps, which hinders their practical applications. Therefore, some recent REC works start to explore other baselines. Based on the end-to-end detection framework DETR~\cite{carion2020end}, Kamath \emph{et.al}~\cite{kamath2021mdetr} propose MDETR, an image vision-language multi-modal pre-training framework benefiting various downstream vision-language tasks. Yang \emph{et.al}~\cite{yang2022tubedetr} propose TubeDETR to conduct spatial-temporal video grounding via a space-time decoder module in a DETR-like manner. However, it still faces some problems: 1) TubeDETR processes each frame independently, which may lead to the loss of temporal information. 2) As a DETR-like method, TubeDETR suffers from slow training convergence. 3) It just fuses visual and language features in a simple concatenation manner and ignores detailed vision-language alignments. In contrast, our ContFormer alleviates the above problems by introducing the content-independent query design and a fine-grained region-phrase alignment.

\noindent \textbf{Transformer Query Design.} DETR~\cite{carion2020end} localizes objects by utilizing learnable queries to probe and filter image regions that contain the target instance. However, this learnable query mechanism has been demonstrated suffering from the slowing training convergence~\cite{liu2022dab,meng2021conditional,wang2021anchor,cao2021pursuit,cao2020task}. To this end, \cite{wang2021anchor} designs object queries based on anchor points to make the queries focus on anchor point areas. \cite{meng2021conditional} proposes a conditional cross-attention mechanism, which attempts to learn the conditional spatial query from decoder embedding and the reference point. \cite{liu2022dab} directly takes box coordinates as the queries and dynamically updates them at each layer. In our work, we employ region-of-interest features as our query design to accelerate convergence.

\noindent \textbf{Vision-Language Alignment.} 
Constructing alignment between visual and language modalities is vital in many vision-language tasks. Most existing methods~\cite{radford2021learning,miech2020end,dong2019dual,xie2020non,cao2019gisca,cao2021unifacegan,cao2021all,yang2021rr} only build coarse-grained alignment (\eg, the image-sentence, video-sentence, region-sentence level alignment), which is not suitable for video REC. Since video REC requires localizing an instance corresponding to representative words, the alignment in video REC should be conducted in the fine-grained region-word alignment. Therefore, in this paper, we contribute two new datasets with region-word annotations and propose to use these labels to regularize the model training and enhance the fine-grained alignment of the query features.

\section{Dataset Illustration}
\begin{figure}[t]
	\centering
	\includegraphics[width=0.5\textwidth]{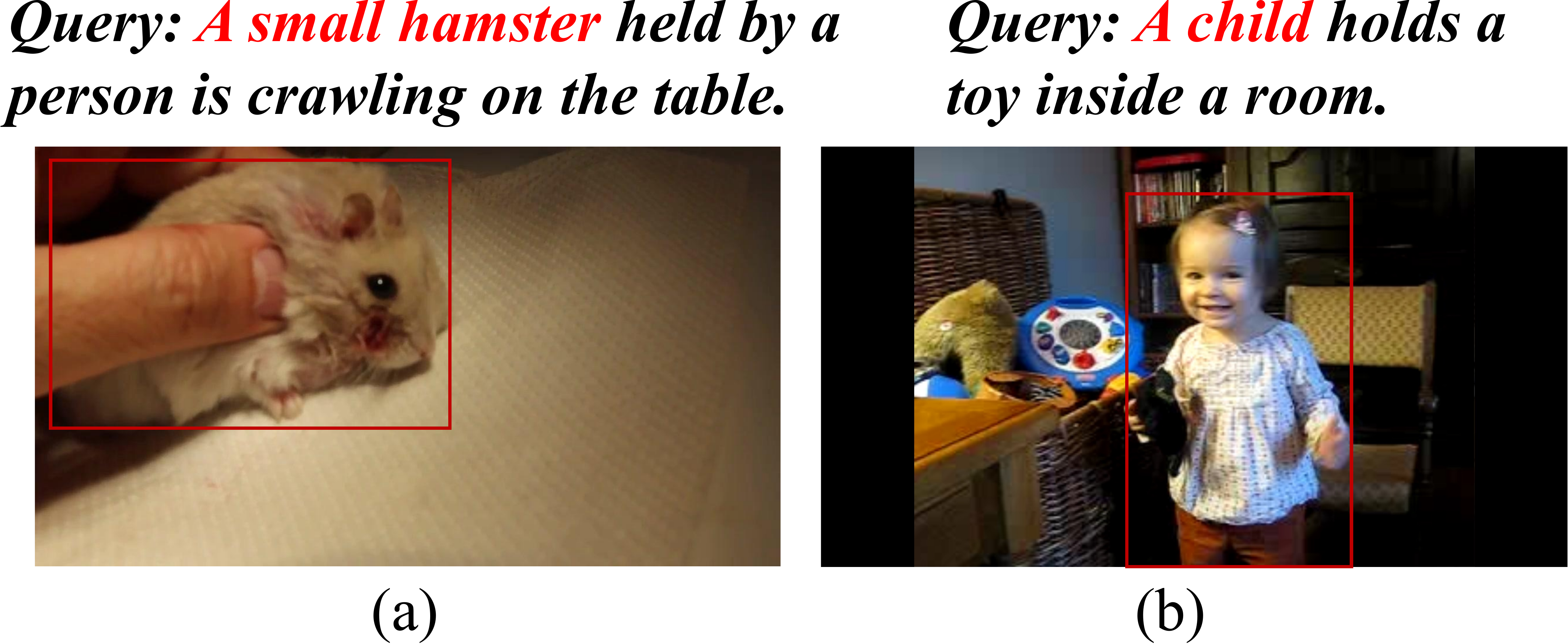}
	\vspace{-1.5em}
	\caption{Visualization examples of (a) VID-Entity and (b) VidSTG-Entity datasets.}
	\label{fig:dataset_example}
\end{figure}

In this section, we give the detailed illustrations of our annotated VID-Entity and VidSTG-Entity datasets. We construct them based on the widely used video REC dataset VID-sentence~\cite{chen2019weakly} and VidSTG~\cite{zhang2020does}. Specifically, VID-sentence and VidSTG contain trimmed and untrimmed videos, respectively. Beyond the existing bounding-box annotations, we explicitly annotate the words corresponding to the region-of-interest. Several examples of our datasets are illustrated in Figure~\ref{fig:dataset_example}.

\noindent\textbf{Dataset Annotation.} To construct fine-grained word-level labels, we utilize Spacy~\cite{honnibal2015improved}, a classic natural language processing model for parsing dependency structure, to indicate the extra positional information of words corresponding to the target. Then, we manually check and correct the annotations. 

\noindent\textbf{Dataset Statistics.}
After annotation, there are 6,582 and 80,684 region-phrase pairs in the training sets of VID-Entity and VidSTG-Entity, respectively. The validation and testing sets of VID-Entity and VidSTG-Entity remain unchanged, which contain  536/536 spatio-temporal tubes with sentences and 8956/10302  video-sentence pairs.

\section{Method}

The schematic illustration of our ContFormer is shown in Figure~\ref{fig:ContFormer_pipeline}. Given the video-language pair as input, our objective is to output a spatial-temporal tube corresponding to the natural language query. Video and language features are extracted by the specific encoder and then are fused by the Transformer encoder (Sec~\ref{sec:3.2}). To ease the efficient training, we propose a Content-aware Query Generation module in the decoder to generate more suitable query features instead of simply setting them learnable. Then the Transformer decoder translates the cross-modal feature into the predicted results (Sec~\ref{sec:3.4}). The overall architecture is optimized by our proposed Entity-aware Contrastive Loss (Sec~\ref{sec:3.5}) to build the fine-grained visual-text alignment.

\subsection{Transformer Encoder}\label{sec:3.2}
The video and text features are extracted by the modality-specific backbone and projected to a shared embedding space, resulting in video features $\boldsymbol{V} \in \mathbb{R}^{T \times C \times H \times W}$ and text features $Y\in \mathbb{R}^{L\times C}$. $T$ is input frame number and $L$ is the word length. $C$ denotes the feature dimension. To ease the Transformer encoder input, we flatten $\boldsymbol{V}$ to generate  $\boldsymbol{U}  \in \mathbb{R}^{F \times C}$, where $F = T \times H \times W$. Then the encoder takes the concatenation of $\boldsymbol{U}$ and $\boldsymbol{Y}$ as input, and generates the cross-modal feature $\boldsymbol{H} \in \mathbb{R}^{ (F + L) \times C}$.
\begin{equation}
	\boldsymbol{H} = \text{MHSA}(\operatorname{Concat}(\boldsymbol{U}, \boldsymbol{Y})),
\end{equation}
\noindent where $\text{MHSA}(\cdot)$ denotes the multi-head self attention. $\operatorname{Concat}(\cdot,\cdot)$ is the concatenation operation.

\subsection{Transformer Decoder.} \label{sec:3.4}

\noindent \textbf{Content-aware Query Generation.}
We propose a novel Content-aware Query Generation module to generate frame content-conditioned query features and thus alleviate the slow training convergence issue. In Figure~\ref{fig:ContFormer_pipeline}, we firstly generate a fix number of learnable regions $\boldsymbol{R} = \left\{\boldsymbol{r}_{0}, \boldsymbol{r}_{1}, ..., \boldsymbol{r}_{N} \right\}$ for each frame and obtain the regional features $\boldsymbol{Q} = \left\{\boldsymbol{q}_{0}, \boldsymbol{q}_{1}, ..., \boldsymbol{q}_{N} \right\}$ by RoI Alignment:
\begin{equation}
	\boldsymbol{q}_{i} = \operatorname{Align}(\boldsymbol{U}, r_i),
\end{equation}
\noindent where $\operatorname{Align}(\cdot, \cdot)$ is the RoI alignment. Note that we omit the frame index since all the frames share the same steps.

\noindent \textbf{Decoder Translation.} Given the visual-language feature $\boldsymbol{H}$ from the encoder output, we directly use the generated regional feature $\boldsymbol{Q}$ as the content-aware query feature. We employ the standard Transformer Decoder to generate the final output $\boldsymbol{P} \in \mathbb{R}^{N \times C}$. 
\begin{equation}
	\boldsymbol{P} = \text{MHCA}(\boldsymbol{H}, \boldsymbol{Q}),
\end{equation}
\noindent where $\text{MHCA}(\cdot,\cdot)$ is the multi-head cross attention in vanilla Transformer decoder. 

To predict the temporal boundary and the bounding box sequence\footnote{Trimmed videos only require bounding box predictions.}, two multi-layer perceptrons (MLP) are applied to generate bounding box predictions $\{\boldsymbol{b}_i\}_{i=1}^{N}$ and temporal predictions $\{\boldsymbol{p}_i\}_{i=1}^{N}$, where $\boldsymbol{b}_i \in \mathbb{R}^{4}$ is the bounding box coordinates and $\boldsymbol{p}_i \in \mathbb{R}^{2}$ represents the frame-wise possibility of the start and end frames.

\subsection{Training loss}\label{sec:3.5}

\noindent\textbf{Bi-partial Matching Loss.} We split the cross-modal feature $\boldsymbol{H} \in \mathbb{R}^{ (F + L) \times C}$ to the attended visual feature $\boldsymbol{H}^{V} \in \mathbb{R}^{ F \times C}$ and text feature $\boldsymbol{H}^{Y} \in \mathbb{R}^{L \times C}$. To find the matching query in our \emph{many-to-one} setting~\cite{cao2021pursuit}, we employ the bi-partial matching strategy. We firstly select the query item with the minimum costs as follows. 
\begin{equation}
	i^* = \underset{i \in [1, N]}{\arg \min}[-\log \boldsymbol{p}_{i} + \mathcal{L}_{\mathrm{box}} + \mathcal{L}_{\mathrm{time}}],
	\label{eq:match1}
\end{equation}
\noindent where $\mathcal{L}_{\mathrm{box}}$ and $\mathcal{L}_{\mathrm{time}}$ are the bounding box regression loss and temporal boundary loss, respectively. 

$\mathcal{L}_{\mathrm{box}}$ is implemented as follows. 
\begin{equation}
	\mathcal{L}_{\mathrm{box}} =\lambda_{\text{giou}}\mathcal{L}_{\text{giou}} +\lambda_{L1}\lvert\lvert \boldsymbol{b}-\hat{\boldsymbol{b}}\rvert\rvert_1,
\end{equation}
\noindent where $\mathcal{L}_{\text{giou}}$ is the scale-invariant generalized intersection over union~\cite{rezatofighi2019generalized}.

$\mathcal{L}_{\mathrm{time}}$ is the temporal boundary loss, which is only used in training untrimmed video datasets (\eg, VidSTG-Entity). We use the Kullback-Leibler divergence values to evaluate the temporal predictions. $\hat{\boldsymbol{b}}$ is the ground truth temporal annotations.
\begin{equation}
	\mathcal{L}_{\text{time}}=\lambda_{\text{KL}}\mathcal{L}_{\text{KL}}(\boldsymbol{b}, \hat{\boldsymbol{b}}).
\end{equation}

The overall matching loss is computed as follows.
\begin{equation}
	\mathcal{L}_{\text{match}} = -\log \boldsymbol{p}_{i^*} + \mathcal{L}_{\mathrm{box}}(\boldsymbol{b}_{i^*}) + \mathcal{L}_{\mathrm{times}}(\boldsymbol{b}_{i^*}).
	\label{eq:match}
\end{equation}

\noindent\textbf{Entity-aware Contrastive Loss.} Benefiting from the fine-grained region-phrase annotations, we build the detailed entity-aware contrastive loss to pull the corresponding word and regional features to be close.
\begin{equation}
	\mathcal{L}_{\text{entity}} = - \log \left(\frac{\exp  (\boldsymbol{H}^V_{i^*})^\text{T} \boldsymbol{H}^Y_{+} / \tau}{\sum^{L}_{k=1} \exp (\boldsymbol{H}^V_{i^*})^\text{T} \boldsymbol{H}^Y_{k}/\tau}\right),
\end{equation}
\noindent where $\boldsymbol{H}^Y_{+}$ is the matched positive sample for $\boldsymbol{H}^V_{i^*}$, $\tau$  is a temperature parameter.

The final loss function is as follows.
\begin{equation}
	\mathcal{L} = \mathcal{L}_{\text{match}} + \lambda_{\text{entity}} \mathcal{L}_{\text{entity}},
\end{equation}
\noindent where $\lambda_{\text{entity}}$ is the balancing factor.

\section{Experiments}

We firstly introduce the experimental settings in Sec.~\ref{sec:5.1}. Then, we compare our ContFormer with the current  state-of-the-art methods in Sec.~\ref{sec:5.2} on both two datasets. In Sec~\ref{sec:5.3}, we further verify the effectiveness of each proposed module. Finally, several visualization results are presented in Sec~\ref{sec:5.4}.
\input{table_fig/VID-entity_sota}

\subsection{Experimental Settings}\label{sec:5.1}

\input{table_fig/vidSTG-entity_sota}
\noindent\textbf{Evaluation Metric.} For the trimmed video dataset, \ie, VID-Entity, we take the commonly used bounding box localization accuracy Accu.@$\eta$ as the metric, where a predicted result is considered correct if the IoU between the predicted region and ground-truth region is greater than a threshold $\eta$. For the untrimmed video dataset, \ie, VidSTG -Entity, we follow~\cite{zhang2020does} to adopt m\_tIoU, m\_vIoU and vIoU@$\theta$ as our evaluation criteria. m\_tIoU is the average temporal IoU between the predicted start-end period and ground-truth start-end period. \cite{zhang2020does} defines vIoU as $\frac{1}{|S_U|}\sum_{t\in S_I}r_t$, where $r_t$ is the IoU between the predicted bounding box and ground-truth bounding box at the $t^{th}$ frame, $S_I$ is the intersection of predicted tubes and ground-truth tubes and $S_U$ is the union of them. m\_vIoU is the average of vIoU and vIoU@$\theta$ is the ratio that vIoU is greater than threshold $\theta$. In this paper, we set $\eta$ to 0.4, 0.5, 0.6 and $\theta$ to 0.3, 0.5.

\noindent\textbf{Implementation details.} For both datasets, we decoded the video by setting \emph{fps} to 5. For the VID-Entity dataset, we set the input frame number $T$ to 20 and the longer edge length to $672$. Since the untrimmed video dataset VidSTG-Entity requires temporal localization, we set $T$ to 200 for the ease of localization. The longer edge length in VidSTG-Entity is set to $224$. We used ResNet101~\cite{he2016deep} pretrained on ImageNet~\cite{deng2009large} as our visual backbone and RoBERTa~\cite{liu2019roberta} pretrained from HuggingFace~\cite{wolf2019huggingface} as the text encoder. We used the AdamW~\cite{loshchilov2017decoupled} optimizer with the initial learning rate setting to $10^{-4}$. The training process lasted for 10 epochs on both datasets. We set $\lambda_{giou}$, $\lambda_{L1}$, $\lambda_{KL}$, $\lambda_{entity}$ and  $\tau$  to 2, 5, 5, 1 and  0.07 respectively.

\input{table_fig/ablation}

\subsection{Comparisons with State-of-the-Arts}\label{sec:5.2}

\noindent\textbf{Results on VID-Entity dataset.} 
We compare our ContFormer with state-of-the-art methods on VID-Entity dataset in Table.~\ref{table:VID-entity}. We classify the compared methods into three categories: \textbf{1) Grounding frame-wisely}. We transfer state-of-the-art image REC methods (\ie, Yang \emph{et.al}~\cite{yang2019fast}, DVSA~\cite{karpathy2015deep}, and GroundeR~\cite{rohrbach2016grounding}) to the video scenario by adding the temporal interaction module (\ie, Avg, NetVLAD~\cite{arandjelovic2016netvlad} and LSTM). 2) \textbf{Tracking frame-wisely}: We attempt to utilize the state-of-the-art tracker~\cite{li2019siamrpn++} to solve video REC. Specifically, we acquire the tracking template on the first/middle/last/random frame by one-stage LSTM~\cite{yang2019fast}, then tracker~\cite{li2019siamrpn++} is applied to track the template according frames.  3) \textbf{Other video REC methods}: We make comparison with state-of-the-art video REC methods including  WSSTG~\cite{chen2019weakly}, and Co-grounding~\cite{song2021co}. As shown in Tabel~\ref{table:VID-entity}, our ContFormer achieves huge improvement boosts. For example, we achieve 7.05\% ,7.02\% and 8.75\% absolute improvement on Accu.@0.4, Accu.@0.5 and Accu.@0.6, respectively. The comparison results demonstrate that our ContFormer has made substantive progress in video REC. 

\input{table_fig/loss_iteration}

\noindent\textbf{Results on VidSTG-Entity dataset.} 
In Table~\ref{table:vidSTG-entity}, we also conduct experiments on the untrimmed video dataset VidSTG-Entity to further explore the effectiveness and generalization of ContFormer. For temporal tube prediction in untrimmed videos, we only utilize a simple temporal boundary loss to realize it, instead of elaborately designing time-aligned cross-attention module in Tubeder~\cite{yang2022tubedetr} or constructing a well-designed 2D temporal feature map in STGRN~\cite{zhang2020learning}. Nonetheless, our ContFormer achieves a greater performance on vIoU@0.3, vIoU@0.5, and m\_vIoU than other state-of-the-art video methods (\ie, STGRN~\cite{zhang2020learning}, STGVT~\cite{su2019vl}, STVGBert~\cite{sharma2018conceptual} and TubeDETR~\cite{yang2022tubedetr}).

\subsection{Ablation studies}\label{sec:5.3}

\noindent\textbf{Content-aware Query Generation.} To investigate the contribution of Content-aware Query Generation(CQG), we train our model with or without CQG. The loss curve comparisons are shown in Figure~\ref{fig:loss_iteration}. Obviously, it can be found that the loss curve converges better with CQG. Moreover, as illustrated in Figure~\ref{fig:accu_epoch}, model without CQG achieves its best performance (Accu.@0.5=65.60\%) at the 9$^{th}$ epoch. In contrast, it only requires 5 epochs to achieve the optimum performance (Accu.@0.5=65.60\%) with CQG. In addition, as shown in Table~\ref{table:ablation}, the performance with CQG has been improved in both VID-Entity and VidSTG-Entity. These experimental results demonstrate the effectiveness of our CQG, which leads to a faster convergence process.

\noindent\textbf{Entity-aware Contrastive Loss.} We further conduct ablation studies of Entity-aware Contrastive Loss (ECL) on VID-entity and VidSTG-entity. Results are summarized in Table~\ref{table:ablation}. On VID-Entity, ECL brings about 4.43\% and 0.053 improvement in Accu.@0.5 and m\_IoU, respectively. In terms of m\_tIoU and m\_vIoU, it achieves 1.96\% and 2.61\% gains on VidSTG-Entity. The significant improvement in spatial localization shows that our ECL successfully guides our model to learn a  better region-word alignment. 

\input{table_fig/K_R}
\input{table_fig/accu_epoch}
\input{table_fig/M_R}
\input{table_fig/visual_examples}

\noindent\textbf{Spatial resolution and temporal length.}  As shown in Table~\ref{table:KR}, we ablate on the values of $T$ and the resolution on the VID-Entity testing. We report the values of Accu.@0.5. According to the experimental results, our model achieves the best performance on VID-Entity when setting $T=20$ and resolution to $672$. As for the ablative results on VidSTG-Entity in Table~\ref{table:MR}, we find the saturated performance is achieved when setting $T=200$  and resolution to $224$.

\subsection{Visualization and Analysis}\label{sec:5.4}

\input{table_fig/query_word}

\noindent\textbf{Region-word Alignment.} We visualize the fine-grained region-word alignment in Figure~\ref{fig:query_word} to further demonstrate the effectiveness of our Entity-aware Contrastive Loss. Specifically, we select some specific object queries and calculate their cosine similarity scores with each word in the described sentence. As shown in Figure~\ref{fig:query_word}, the query containing the target instance "\texttt{cat}" is highly corresponding to word "\texttt{cat}", which manifests our motivation that Entity-aware Contrastive Loss leads to more fine-grained alignment. More visualizations of region-word alignment are provided in the supplementary. 

\noindent\textbf{Grounding results.}
We provide the visualization results of our ContFormer and other state-of-the-art methods in Figure~\ref{fig:visual_examples}. Results demonstrate the superiority of our method more intuitively. Specifically, in Figure~\ref{fig:visual_examples}(a), our ContFormer localizes the region of "\texttt{squirrel}" precisely, while \cite{song2021co} fails to ground the target instance due to the interference of occlusion. In Figure~\ref{fig:visual_examples}(c), the target instance "\texttt{rabbit}" is too small and the word "\texttt{deer}" in the query sentence may confuse model to localize the non-target instance. Nevertheless, our ContFormer still captures the overall semantic information and targets the corresponding instance in spatial region correctly. Similar to Figure~\ref{fig:visual_examples}(c), the example of VidSTG-Entity faces the same challenge in grounding target. However, ContFormer both localizes the target region precisely, which shows the effectiveness of our Content-aware Query Generation and Entity-aware Contrastive Loss. Fine-grained alignment visualization of these examples and more grounding results can be found in the supplementary material.

\section{Conclusion}
We proposed ContFormer, a novel Transformer-based video REC method with the context-aware query feature. Specifically, instead of using purely learnable embeddings without any explicit physical meanings, we designed a content-aware query generation module to generate content-conditioned queries, which lead to a better convergence process. Besides, we propose an entity-aware contrastive loss to construct the fine-grained visual-text alignment for our ContFormer. For the convenience of this, we contributed two video datasets (\ie, VID-Entity and VidSTG-Entity) which contain the region-phrase alignment annotations. Extensive experimental results show that our ContFormer achieves state-of-the-art performance on both trimmed and untrimmed video REC datasets.

{\small
\bibliographystyle{ieee_fullname}
\bibliography{egbib}
}

\end{document}

%% file: table_fig/VID-entity_sota.tex
\begin{table}[h]
\caption{Comparison (\%) with state-of-the-art methods on VID-Entity dataset.}
\vspace{-0.5em}
\renewcommand\arraystretch{1.1}
\label{table:VID-entity}
\begin{center}
\resizebox{\linewidth}{!}{
\begin{tabular}{ccccccccccccccc}
\toprule
\multirow{2}*{\textbf{Method}} & \multicolumn{3}{c}{\textbf{Accu.@}} \\
\cmidrule{2-4}
& 0.4 & 0.5 & 0.6  \\
\midrule
Yang \emph{et al.} (\emph{w/} BERT) & - & 52.39 & -  \\
Yang \emph{et al.} (\emph{w/} LSTM) & - & 54.78 & -  \\
DVSA +Avg & 36.2 & 29.7 & 23.5 \\
DVSA +NetVLAD & 31.2 & 24.8 & 18.5  \\
DVSA +LSTM & 38.2  & 31.2 & 23.5  \\
GroundeR +Avg & 36.7 & 31.9 & 25.0 \\
GroundeR+NetVLAD & 26.1 & 22.2 & 15.1  \\  
GroundeR +LSTM & 36.8 & 31.2 & 27.1  \\
\cdashline{1-4}[2pt/2pt]
First-frame tracking & - & 36.97 & -  \\
Middle-frame tracking & - & 44.00 & - \\
Last-frame tracking & - & 36.26 & -  \\
Random-frame tracking & - & 40.20 & -\\
\cdashline{1-4}[2pt/2pt]
WSSTG & 44.60 & 38.20 & 28.90 \\
Co-grounding & 63.35  & 60.25 & 53.89  \\
\cdashline{1-6}[2pt/2pt]
\textbf{ContFormer (Ours)} & \textbf{70.40} & \textbf{67.27} & \textbf{62.64} \\
\bottomrule
\end{tabular}
}
\end{center}
\end{table}

%% file: table_fig/vidSTG-entity_sota.tex
\begin{table*}[t]
\caption{Comparison (\%) with state-of-the-art methods on VidSTG-Entity dataset. (VG: Visual Genome~\cite{krishna2017visual}, CC: Conceptual Captions~\cite{sharma2018conceptual}, IN: ImageNet~\cite{deng2009imagenet}). PT data denotes pretraining data.}
\vspace{-0.5em}
\renewcommand\arraystretch{1.1}
\setlength{\tabcolsep}{5pt}
\label{table:vidSTG-entity}
\begin{center}
\resizebox{\linewidth}{!}{
\begin{tabular}{lcccccccccccccccccccccc}
\toprule
\multirow{2}*{\textbf{Method}} &\multirow{2}*{\textbf{PT Data}}&& \multicolumn{4}{c}{\textbf{Declarative Sentences}}&& \multicolumn{4}{c}{\textbf{Interrogative  Sentences}} \\
\cmidrule{4-7}
\cmidrule{9-12}
&&& m\_tIoU& m\_vIoU & vIoU@0.3 & vIoU@0.5&& m\_tIoU& m\_vIoU & vIoU@0.3 & vIoU@0.5 \\
\midrule
STGRN & VG &&\textbf {48.5} &19.8 & 25.8 & 14.6 && \textbf{47.0} & 18.3 & 21.1 &12.8 \\
STGVT & VG+CC && - &  21.6  & 29.8 & 18.9 && - & - & - & - \\
STVGBert & IN+VG+CC && - &  24.0  &  30.9 &  18.4  && - &  22.5 & 26.0 & 16.0\\
TubeDETR & IN && 43.1 &  28.0  &  39.9 &  26.6  && 42.3 &  25.1 & 35.7 & 22.4\\
\textbf{ContFormer (Ours)} & IN && 44.9 &  \textbf{29.7}  &  \textbf{40.1 }& \textbf{ 27.8}  && 43.5 & \textbf{ 26.4} & \textbf{36.0} & \textbf{23.5} \\
\bottomrule
\end{tabular}
}
\end{center}
\end{table*}

%% file: table_fig/ablation.tex
\begin{table}[t] 
\caption{Video REC results for ablation study on trimmed video dataset VID-Entity and untrimmed video dataset VidSTG-Entity, respectively. (CQG: Content-aware Query Generation, ECL: Entity-aware Constrastive Loss)}
\vspace{-0.5em}
\renewcommand\arraystretch{1.1}
\setlength{\tabcolsep}{2pt}
\label{table:ablation}
\begin{center}
\resizebox{\linewidth}{!}{
\begin{tabular}{lcccccccccccccc}
\toprule
\multirow{2}*{\textbf{Mode}} &\multirow{2}*{\textbf{CQG}}&\multirow{2}*{\textbf{ECL}} &\multicolumn{2}{c}{\textbf{VID-Entity}} && \multicolumn{2}{c}{\textbf{VidSTG-Entity}}\\
\cmidrule{4-5}
\cmidrule{7-8}

&&&Accu.@0.5& m\_IoU&&m\_tIoU&m\_vIoU  \\
\midrule
\#1&\ding{51} & \ding{51} & \textbf{67.17}&\textbf{0.600}&&\textbf{44.21}&\textbf{28.08}\\
\#2&\ding{55} & \ding{51} & 65.60&0.596&&43.01&27.38\\
\#3&\ding{51} & \ding{55} & 62.10&0.551&&41.72&25.56\\
\#4&\ding{55} & \ding{55} & 61.17&0.543&&41.05&24.77\\

\bottomrule
\end{tabular}
}
\end{center}
\end{table}

%% file: table_fig/loss_iteration.tex
\begin{figure}[t]
	\centering

	\includegraphics[width=0.4\textwidth]{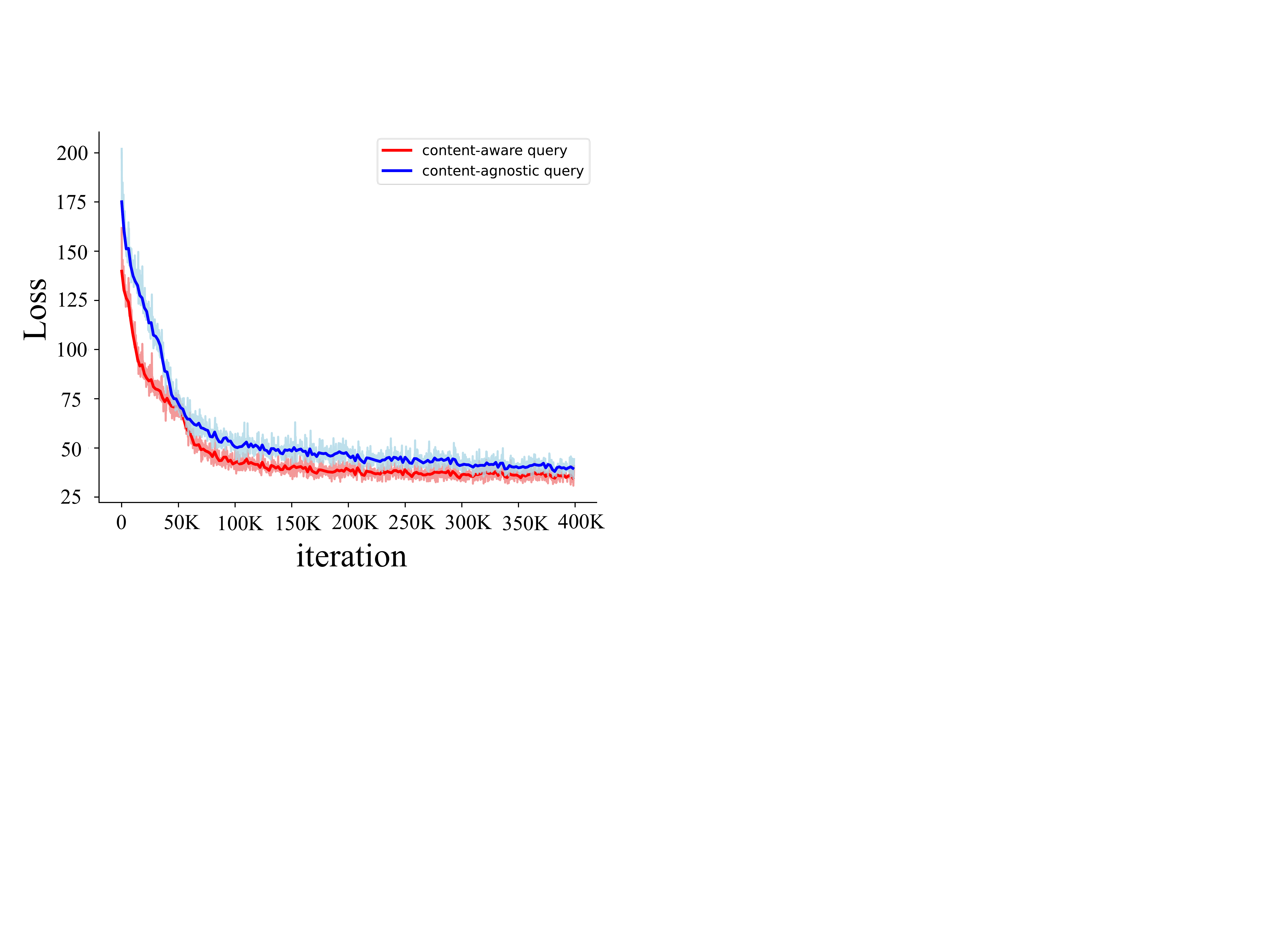}
	\vspace{-0.5em}
	\caption{Comparison of loss curve of content-aware query (marked in \textcolor{red}{red}) and content-agnostic query (marked in \textcolor{blue}{blue}).}
	\label{fig:loss_iteration}
\end{figure}

%% file: table_fig/K_R.tex
\begin{table}[t]
\vspace{-0.5em}
\caption{Comparison of performance with various $T$ and spatial resolution on the VID-Entity testing set. }
\vspace{-0.5em}
\renewcommand\arraystretch{1.1}
\label{table:KR}
\begin{center}
\begin{tabular}{ccc}
\toprule
\textbf{$T$} & \textbf{Resolution} & \textbf{Accu.@0.5} \\
\midrule
15  & 800 & 66.38  \\
20 & 672 &\textbf{67.17}\\
25 & 576 &65.93\\
35 & 480 &59.28\\
\bottomrule
\end{tabular}
\end{center}
\end{table}

%% file: table_fig/accu_epoch.tex
\begin{figure}[t]
	\centering
	\includegraphics[width=0.4\textwidth]{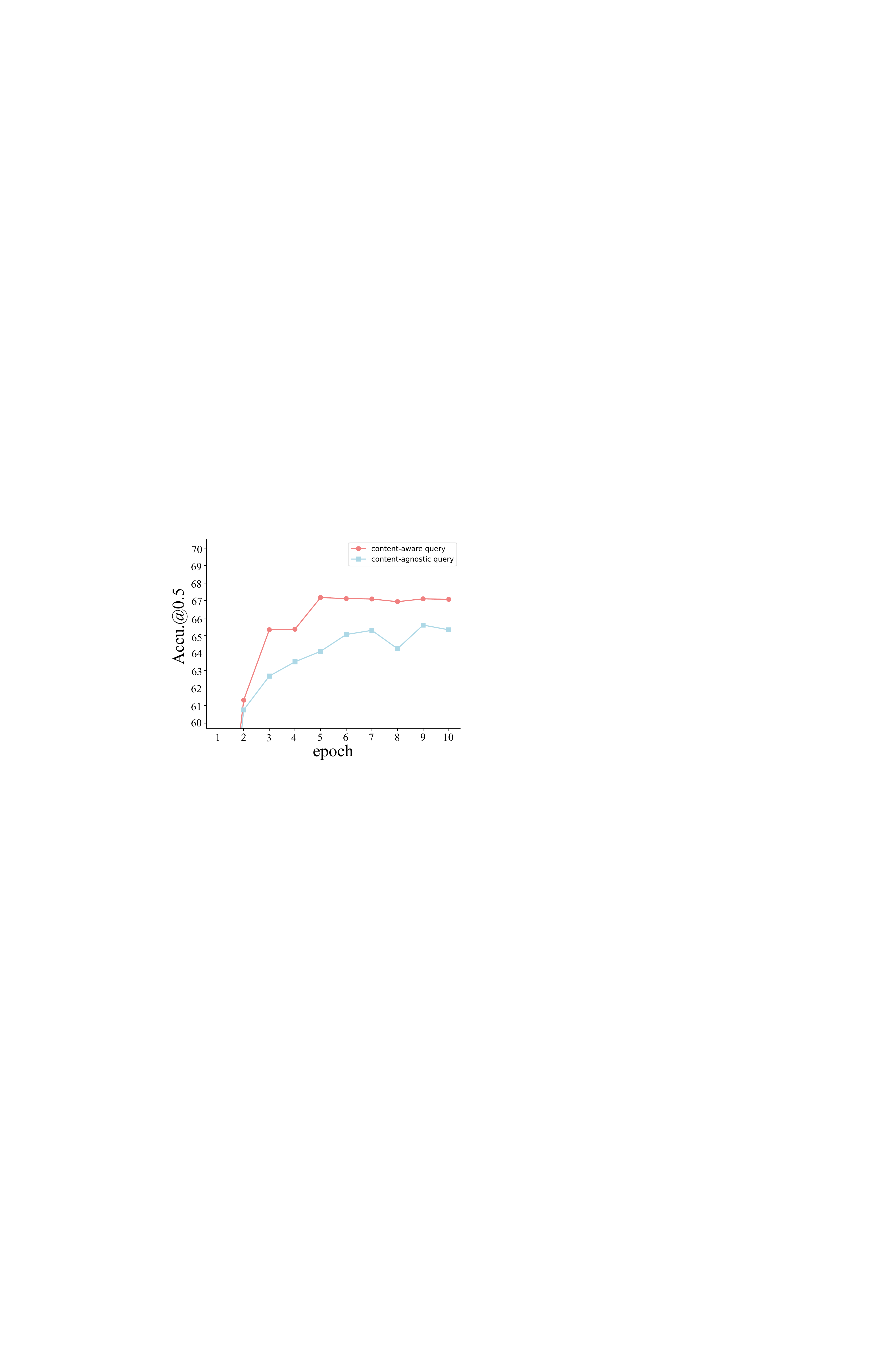}
	\vspace{-0.5em}
	\caption{Comparison of Accu.@0.5 between content-aware query (marked in \textcolor{red}{red}) and content-agnostic query (marked in \textcolor{blue}{blue}).}
	\label{fig:accu_epoch}
\end{figure}

%% file: table_fig/M_R.tex
\begin{table}[t]
\vspace{-0.5em}
\caption{Comparison of performance with various maximum of $T$ and spatial resolution on the VidSTG-Entity testing set. }
\vspace{-0.5em}
\renewcommand\arraystretch{1.1}
\label{table:MR}
\begin{center}
\begin{tabular}{cccc}
\toprule
\textbf{$T$} & \textbf{Resolution} & \textbf{m\_tIoU} & \textbf{m\_vIoU} \\
\midrule
 100 & 288&42.16&27.31 \\
 150 & 256 &43.33&27.56\\
 200 & 224 &\textbf{44.21}&\textbf{28.08}\\
\bottomrule
\end{tabular}
\end{center}
\end{table}

%% file: table_fig/visual_examples.tex
\begin{figure*}[t]
	\centering
	\includegraphics[width=0.85\textwidth]{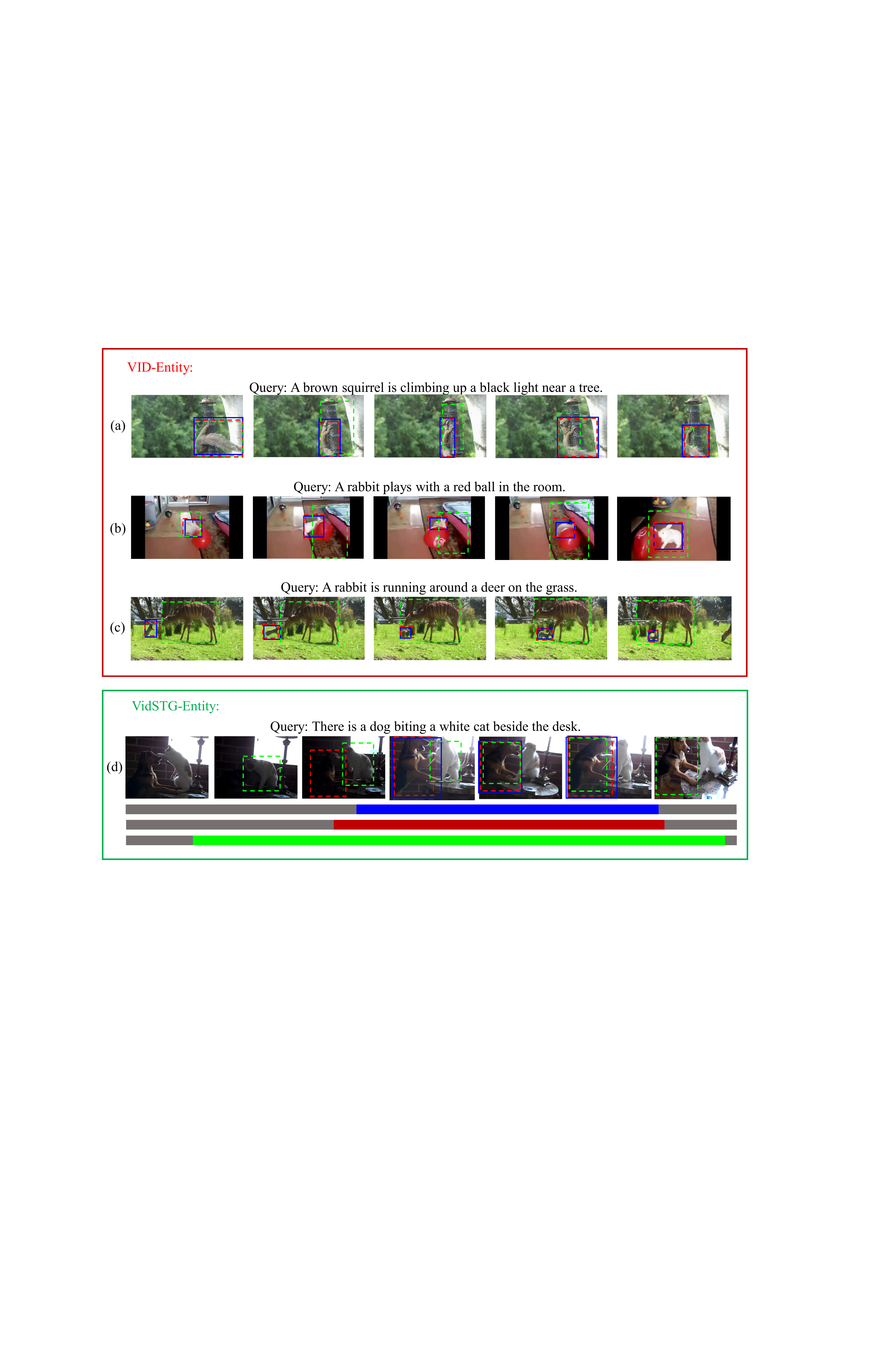}
	\vspace{-0.5em}
	\caption{Qualitative results of our ContFormer (marked in \textcolor{red}{red} dotted box), compared with \cite{song2021co} in VID-Entity and \cite{yang2022tubedetr} in VidSTG-Entity. The compared methods are marked in \textcolor{green}{green} dotted box. The ground-truth annotations are marked in \textcolor{blue}{blue}.}
	\label{fig:visual_examples}
\end{figure*}

%% file: table_fig/query_word.tex
\begin{figure}[h]
	\centering
	\includegraphics[width=0.4\textwidth]{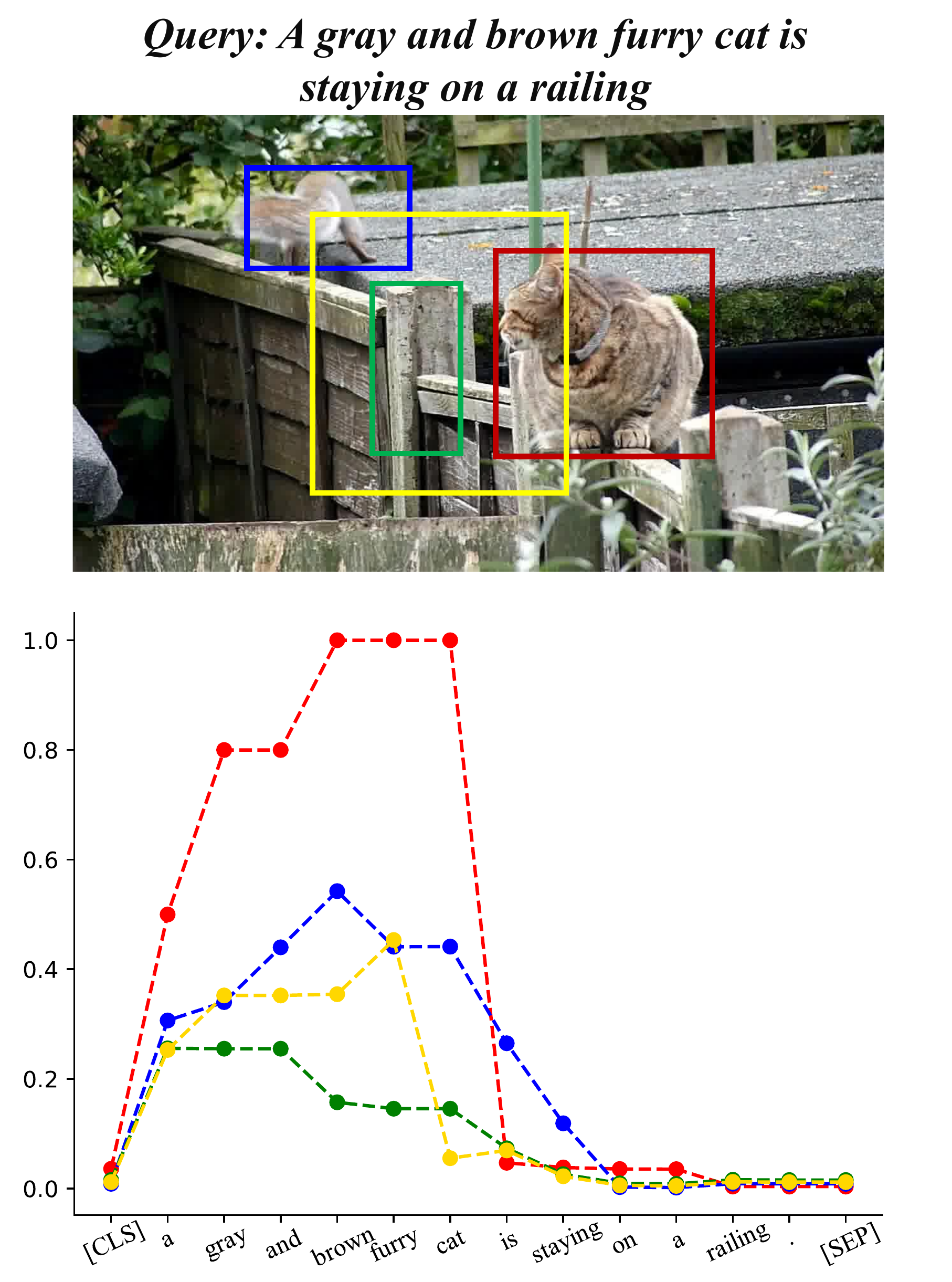}
	\vspace{-0.5em}
	\caption{Illustration of fine-grained region-word level alignment. The regions marked by bounding boxes are corresponding to dashed lines according to color.}
	\label{fig:query_word}
\end{figure}